\documentclass[conference]{IEEEtran}
\usepackage{times}
\usepackage{multicol}
\usepackage[bookmarks=true]{hyperref}

\usepackage{lipsum}
\usepackage{xcolor}
\usepackage{amssymb}
\usepackage{graphicx}
\usepackage{subcaption}
\usepackage{amsmath}
\usepackage{bbm}
\usepackage{amsfonts}
\usepackage{xfrac}
\usepackage{float}
\usepackage{booktabs,tabularx}
\usepackage{array}
\newtheorem{problem}{Problem}

\usepackage[%
    style=numeric-comp,
    sorting=none,
    natbib=true,  %
    backend=biber,
    sortcites=true,
    doi=false,
    url=false,
    giveninits=true,
    mincitenames=1,
    maxcitenames=1, %
    minbibnames=3, %
    maxbibnames=4, %
    hyperref
]{biblatex}
\bibliography{bibtex/external, bibtex/Swarmlab}

\newcommand{\x}{x} %
\newcommand{\y}{y} %
\newcommand{\cont}{c} %
\newcommand{\modal}{m} %
\newcommand{\reward}{r} %
\newcommand{\price}{p} %
\newcommand{\cluster}{\mathcal{C}} %
\newcommand{\suc}{p} %
\newcommand{\Ncluster}{N} %
\newcommand{\propor}{\rho} %
\newcommand{\algo}{\boldsymbol{A}}

\newcommand{\refeq}[1]{(\ref{#1})}
\newcommand{\reffig}[1]{Fig. \ref{#1}}
\newcommand{\refapp}[1]{App. \ref{#1}}

\newcommand{\refprob}[1]{Problem \ref{#1}}

\begin{document}

\title{Online Foundation Model Selection in Robotics}

\author{
    Po-han Li$^1$, Oyku Selin Toprak$^2$, Aditya Narayanan$^1$, Ufuk Topcu$^1$, Sandeep Chinchali$^1$\\
    \authorblockA{$^1$The University of Texas at Austin, $^2$Bogazici University}
}

\maketitle

\begin{abstract}
Foundation models have recently expanded into robotics after excelling in computer vision and natural language processing. The models are accessible in two ways: open-source or paid, closed-source options. Users with access to both face a problem when deciding between effective yet costly closed-source models and free but less powerful open-source alternatives. We call it the \textit{model selection problem}. Existing supervised-learning methods are impractical due to the high cost of collecting extensive training data from closed-source models. Hence, we focus on the online learning setting where algorithms learn while collecting data, eliminating the need for large pre-collected datasets. We thus formulate a user-centric \textit{online model selection problem} and propose a novel solution that combines an open-source encoder to output context and an online learning algorithm that processes this context. The encoder distills vast data distributions into low-dimensional features, \textit{i.e.}, the \textit{context}, without additional training. The online learning algorithm aims to maximize a composite reward that includes model performance, execution time, and costs based on the context extracted from the data. It results in an improved trade-off between selecting open-source and closed-source models compared to non-contextual methods, as validated by our theoretical analysis. Experiments across language-based robotic tasks such as Waymo Open Dataset, ALFRED, and Open X-Embodiment demonstrate real-world applications of the solution. The results show that the solution significantly improves the task success rate by up to $14\%$.

\end{abstract}
\IEEEpeerreviewmaketitle

\section{Introduction}
\begin{figure}[t!]
\centering
\includegraphics[width=0.5\textwidth]{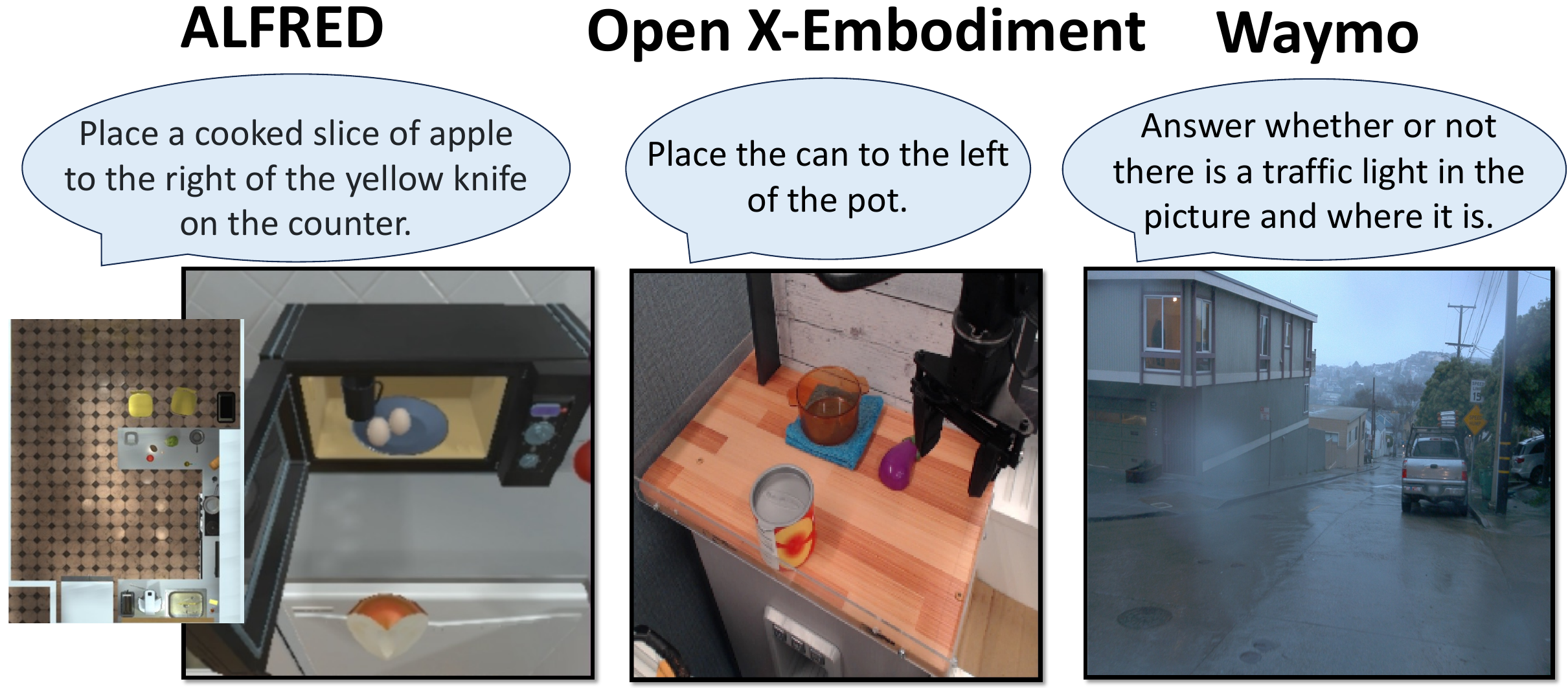}
\caption{\small{\textbf{Examples of language-based robotic tasks:} 
    Robots receive language instructions from users and visual observations from the environment and output either actions to pilot the robots (ALFRED and Open X-Embodiment) or answers to the users (Waymo).
}}
\label{fig:datasets}
\vspace{-2em}
\end{figure}

Foundation models, excelling in computer vision \cite{kirillov2023segmentanything,oquab2023dinov2,simCLRv2} and natural language processing \cite{openai2023gpt4,touvron2023llama,touvron2023llama2}, have recently expanded into robotics.
These models exhibit the ability to handle various tasks and operate multiple robotic embodiments through language instructions \cite{RTX, saycan, huang2022language, firoozi2023foundation}.
\reffig{fig:datasets} shows examples of such tasks.
Some foundation models are open-source and publicly accessible, such as the Octo models \cite{octo_2023} trained on the Open X-Embodiment Dataset \cite{RTX}.
Others are closed-source, keeping their training code, architecture, and weights confidential, such as Google's RT-X model \cite{RTX} trained on the same dataset.
Closed-source models often operate as paid services, which charge users for each access and are usually more powerful than open-source models.
However, their opacity restricts the users from knowing the models' biases and limitations.
The trade-off between powerful but costly closed-source models and their open-source counterparts poses a \textit{model selection problem} for users with access to both types of models.
The users must determine the most suitable model for their specific needs from all open-source and closed-source options, each with unique attributes.

\begin{figure*}[t!]
  \centering
  \includegraphics[width=0.85\textwidth]{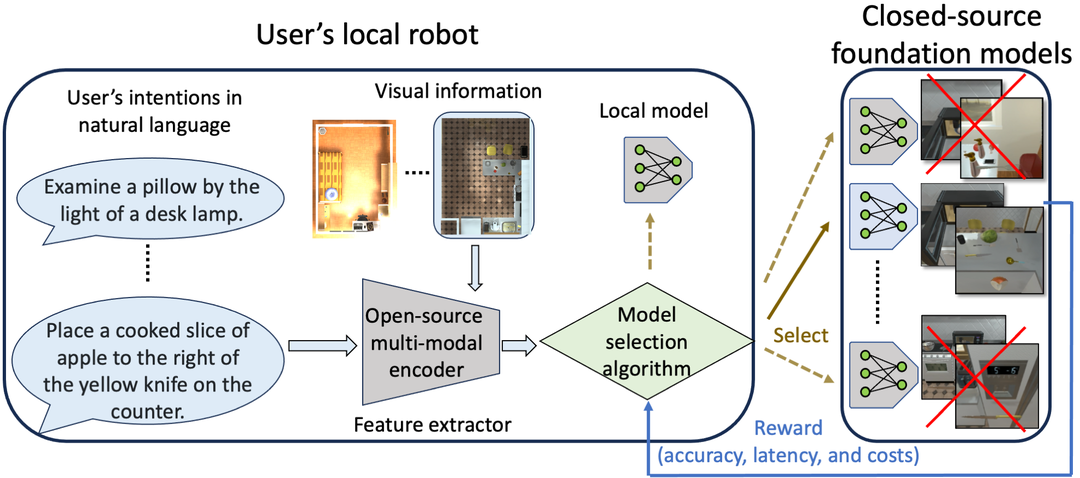}
  \caption{\small{\textbf{Online model selection pipeline:}
    A user sends their intentions in natural language and images to a model selected from a range of available options.
    To do so, an encoder first processes the language and visual inputs to extract features. 
    These features help an online learning algorithm select the suitable model that maximizes accuracy and minimizes response latency and monetary costs. 
    The algorithm should avoid selecting models that execute incorrectly, marked with red crosses. 
    The above examples come from the ALFRED dataset.
    }}
  \label{fig:system_graph}
\vspace{-2em}
\end{figure*}

While prior works have suggested the use of supervised-learning methods for model selection problems \cite{chen2023frugalgpt,frugalml,efficientML,ghosh2022modelselectionRL,chinchali2021network}, the extensive training data required make it costly and time-consuming with closed-source models.
The challenge is particularly acute for individual users with limited resources compared to larger corporations.
We thus focus on the online learning setting where algorithms simultaneously learn and collect data, thereby removing the necessity for extensive pre-collected datasets.
We then formulate the model selection problem with closed-source models as an online learning problem, called the \textit{online model selection problem}.
We provide a solution to this problem that uses limited data to explore various models efficiently.

Given the observation that collecting training data for closed-source models is expensive, our insight is to leverage a pre-trained, open-source encoder \cite{clip, ViT, devlin2019bert, siglip} to accelerate learning from limited data. 
This encoder trained on extensive datasets compresses data into lower dimensions featuring inherent clustering structures, simplifying the learning task for online model selection algorithms and thus accelerating learning. 
Also, it can handle various data distributions and work as a zero-shot feature extractor.
We consider scenarios where users can select an open-source foundation model deployed on their local devices instead of remote, closed-source foundation models. 
The closed-source models run remotely on private servers since they are not publicly accessible, and the local model allows users free access without the need to send data to distant servers.
Although the local model may not match the performance of its remote counterparts, it offers a cost and time-efficient alternative for users, particularly for easy tasks.

We present a user-centric online learning pipeline to efficiently select a local, open-source model or remote, closed-source model, as shown in \reffig{fig:system_graph}.
First, a user who has access to both types of models expresses their intention through natural language and an accompanying image, such as a query of ``place a cooked slice of apple on the counter" and a floor plan image of the kitchen. 
An open-source encoder processes these inputs into fixed-dimensional features, which we call the context. 
A contextual online learning algorithm selects the optimal model, local or remote, by analyzing the specific context of each data input.
The selected model takes the intention sentence and the image as input and generates a final output, such as the actions that pilot a household robot to execute the user's intended task.
Finally, a reward is obtained based on the model execution. It encompasses model performance, model execution time, and monetary cost, representing the user experience, and only the selected model returns the reward.
The online model selection algorithm continually learns to maximize this reward, thus improving the user experience. 
Our pipeline can easily adapt to new model releases by updating the selection model algorithm to select from more models without the need to retrain the algorithm from scratch.
It is crucial given the frequent publication of foundation models.
\reffig{fig:system_graph} illustrates an example where the encoder processes two modalities, vision and language, typical in language-based robotic tasks. However, note that our pipeline is capable of processing single or more modalities.

\textbf{Contributions:}
Our contributions are threefold:
First, we formulate the model selection problem with closed-source models as an online model selection problem.
Second, we propose a novel solution to effectively solve the formulated problem, which combines an open-source encoder and a contextual online learning algorithm.
Lastly, our theoretical analysis characterizes the trade-off between selecting local and remote models.
We also theoretically characterize the benefit of using contextual algorithms rather than non-contextual ones.
We validate our proposed methods with a question-answering task, Massive Multitask Language Understanding (MMLU) \cite{hendryckstest2021mmlu}, and three language-based robotic tasks, Waymo Open Dataset \cite{Sun_2020_Waymo_dataset}, ALFRED \cite{ALFRED20}, and Open X-Embodiment \cite{RTX}.
Our results show an improvement in the task success rate of up to $14\%$ compared to non-contextual methods.

\section{Related Work}
\textbf{Foundation models:}
Foundation models, pre-trained on extensive data, offer wide-ranging capabilities across multiple domains.
In language processing, open-source foundation models, such as BERT \cite{devlin2019bert}, Vicuna \cite{vicuna}, and LLaMA \cite{touvron2023llama, touvron2023llama2} are capable of most language-based tasks, such as answering questions and sentiment analysis. 
In computer vision, models such as SAM \cite{kirillov2023segmentanything}, DINOv2 \cite{oquab2023dinov2}, and SegGPT \cite{wang2023seggpt} demonstrate competitive performance without task-specific training, which excel in segmenting a wide range of images. 
In robotics, open-source models leverage insights from large language models to execute interactive tasks, demonstrating their ability to break down high-level tasks into mid-level plans \cite{huang2022language, octo_2023}.
Closed-source models also show multi-modal capabilities and are more powerful. Closed-source large language models, such as GPT-4 \cite{openai2023gpt4}, Bard \cite{Bard}, and Gemini \cite{gemini}, can process user prompts along with images and generate text outputs that are indistinguishable from human-generated text.
In robotics, foundation models interpret various data modalities. These closed-source models \cite{saycan, RTX} facilitate interactive conversations and control robots using multi-modal inputs, such as vision and language, which allows them to adapt to changing enviroments without the need for pre-defining every scenario.

\textbf{Model selection algorithms:}
Existing works on model selection problems use supervised-learning methods across various domains, such as computer vision \cite{frugalml, efficientML}, natural language processing \cite{chen2023frugalgpt}, drug design \cite{drug_bandit}, robotics \cite{chinchali2021network, ghosh2022dynamic}, and reinforcement learning \cite{ghosh2022modelselectionRL, singi2023decisionRL, osa2023offline}. 
These methods require extensive training data, while our approach prioritizes data efficiency in the online learning setting.
The most similar work to ours learns an online learning algorithm to combine the outputs of all models \cite{karimi2021onlineactive}. It differs from ours since selecting all closed-source models is costly and impractical.

\textbf{Contextual online learning:}
Previous works on contextual online learning algorithms mainly focus on settings where context lies in a compact set, and they provide provable performance guarantees for these algorithms \cite{ContextualMulti-ArmedBandits, turgay2018contextual, slivkins2011contextual}.
However, the output range of our extracted features is unbounded and thus not suitable for these methods.
Other algorithms dynamically form clusters and then train a non-contextual learning algorithm for each cluster \cite{gentile2014online, wang2023adcb, nguyen2014dynamic}. 
They fall short in processing high-dimensional data, prompting us to modify deep on-policy reinforcement learning algorithms for our setting.

\section{Problem Formulation}

We formulate the online model selection problem as a contextual online learning problem.
It involves a contextual algorithm that continually learns from a sequence of data, represented by the context, and the corresponding rewards from the algorithm's outputs.
In this problem, a user selects a model from a set $\mathcal{F}$ of models to execute and meet their specific requirements.
We denote the set $\mathcal{F}$ of $k+1$ models as $\{\boldsymbol{f}^0,\boldsymbol{f}^1,...,\boldsymbol{f}^k\}$, where $\boldsymbol{f^0}$ is the local, open-source foundation model, and the others are remote, closed-source ones. 
At each time step $t$, the user observes a data point $\x_t=(\x_t^1, \x_t^2, ..., \x_t^\modal)$, which is sampled from an unknown, $\modal$-modal, independent and identically distributed distribution $\mathcal{X}$.
We denote a sequence of data from time $1$ to $T$ as $\x_{1:T}$. 
Each data point $\x_t$ has an inaccessible label $\y_t$ that we aim to infer, and each model $\boldsymbol{f}^i$ can process a data point $\x_t$ and outputs an inferred label $\hat{\y}_t^i = \boldsymbol{f}^i(\x_t)$.
For example, the data point $\x_t$ can be the pair of user instructions and current observations of a robot, and the label $\y_t$ can be the optimal actions of a robot to complete the user's requests.

The user employs a contextual online learning algorithm $\algo$ to learn to select the model. This algorithm is a function that takes the context $\cont_t$ of the data $\x_t$ as input and outputs an action $a_t$.
The action $a_t$ is an integer between $0$ and $k$, representing the selected model to input data point $x_t$ from the set $\mathcal{F}$ at each time step $t$.
We use an encoder $\boldsymbol{g}$ to extract features, \textit{i.e.}, the context, from the data and denote the extracted context from data point $x_t$ as $\cont_t$, namely,
$$\cont_t=\boldsymbol{g}(\x_t).$$
The encoder can contain various parts encoding different modalities of data points. For simplicity, we use $\boldsymbol{g}$ to denote the whole encoder that inputs all modalities.

Finally, the algorithm $\algo$ receives a reward $\reward_t$ that reflects the quality of the inferred label $\hat{\y}_t$, the total execution time of the model, and monetary cost:

    \begin{equation}
    \reward_t = \underbrace{\boldsymbol{S}(\y_t, \hat{\y}_t)}_{\substack{\text{model} \\ \text{performance}}} - 
    \alpha_{\tau}
    \underbrace{\boldsymbol{\tau}(x_t,f^{a_{t}})}_{\substack{\text{execution} \\ \text{time latency}}} - 
    \alpha_{\price}
    \underbrace{\boldsymbol{\price}(x_t,f^{a_{t}})}_{\text{monetary cost}}.
    \label{eq:reward}
\end{equation}
The model performance $\boldsymbol{S}$ measures the quality of the inferred label by assessing its similarity to the ground truth labels, such as the mean-squared error between the model's output actions and the optimal actions. 
It can also indicate whether the robot successfully executed the tasks or not.
The action $a_t$ at time $t$ indicates the selected model $\boldsymbol{f^{a_{t}}}$.
Time function $\boldsymbol{\tau}$ takes the data point and the selected model as inputs and outputs the execution time, while price function $\boldsymbol{\price}$ takes the same two terms as inputs and outputs the price of the execution.
Non-negative weights $\alpha_{\tau}$ and $\alpha_{\price}$ balance the execution time and the price.
Typically, the precise outputs of the time and price function cannot be predetermined due to the unpredictable nature of the models' outputs.
Hence, the algorithm $\algo$ must learn to predict the output of the time function $\boldsymbol{\tau}$ and the price function $\boldsymbol{\price}$.
The objective is to maximize the cumulative mean rewards $\frac{1}{T}\sum_{t=1}^T r_t$, 
\textit{i.e.}, maximizing the model performance and minimizing the execution time and costs.

We now define the online model selection problem, where the algorithm $\algo$ maximizes the cumulative mean of the expected rewards due to the inherent stochasticity of the input data distribution $\mathcal{X}$ and 
models $\{\boldsymbol{f}^0,\boldsymbol{f}^1,...,\boldsymbol{f}^k\}$.
\begin{problem}[Online Model Selection Problem] 
\label{prob}
Given a sequence of multi-modal data $\x_{1:T}$, an encoder $\boldsymbol{g}$, and a set $\mathcal{F}$ of models $\{\boldsymbol{f^0},...,\boldsymbol{f^k}\}$, we aim to find the optimal algorithm $\algo$ that maximizes the cumulative mean of the expected rewards:
\begin{equation*}
\begin{aligned}
    \max_{\algo} & ~ \frac{1}{T} \mathbb{E}_{\mathcal{X}} \left[\sum_{t=1}^T \mathbb{E}_{f^{a_t}} \left[ \reward_t \right] \right] \\
    \text{s.t. } 
    & \cont_t = \boldsymbol{g}(\x_t) \\
    & a_t = \algo(\cont_t) \\
    & \hat{y}_t^{a_t} = \boldsymbol{f}^{a_t}(x_t) \\ 
    & \reward_t = \boldsymbol{S}(\y_t, \hat{\y}_t) - \alpha_{\tau} \boldsymbol{\tau}(x_t;f^{a_{t}}) - \alpha_{\price} \boldsymbol{\price}(x_t;f^{a_{t}}).
\end{aligned}
\end{equation*}
\end{problem}
Note that the expectation of the reward is on the selected model $\boldsymbol{f^{a_t}}$ and the data distribution $\mathcal{X}$, and the algorithm takes the context $\cont_t$ instead of the raw data $\x_t$ as input. 
For the rest of the paper, we extend the horizon $T$ to infinity, denoted as $T \rightarrow \infty$. It changes the focus of the online learning process from maximizing immediate rewards to the long-term average of rewards.
The purpose of this horizon shift is to improve overall performance by reducing the impact of short-term random variations.
we compare baseline algorithms by their cumulative mean rewards for every current time step $T_{\mathrm{cur}}$:
\begin{equation}
R_{\mathrm{cum}} = \frac{1}{T_{\mathrm{cur}}}\sum_{t=1}^{T_{\mathrm{cur}}} r_t \quad (\text{current cumulative~mean~rewards}).
\label{eq:cur_reward}
\end{equation}
Here, higher cumulative mean rewards indicate greater effectiveness of a certain algorithm over the evaluated period $T_{\mathrm{cur}}$.

\section{Online Model Selection}

We now provide a detailed description of our online model selection pipeline, as illustrated in \reffig{fig:system_graph}. 
A user observes a stream of data $\x_{1:T}$ and selects a model from $\mathcal{F}$ to execute.
For example, a user expresses their intentions to a home robot.
They can say ``place a cooked slice of apple on the counter" and provide the robot with the floor plan of a kitchen.
The household robot with limited computing power and a less performant local open-source model can choose a closed-source model to control itself, incurring costs. 
A contextual online model selection algorithm on the robot continuously learns to maximize a composite reward, including model performance, execution time, and monetary costs.

\textbf{Input query data:} 
The input data stream $\x_{1:T}$ from time $1$ to $T$ constitutes $\modal$-modalalities that the user observes visually or describes using natural language. 
Determining the user's intended tasks beforehand is difficult since they could encompass vast data distributions and tasks.
For example, the user can tell the household robot to prepare meals, organize items, and interact with any object in the room, as exemplified in the ALFRED dataset \cite{ALFRED20} in \reffig{fig:datasets} and \reffig{fig:system_graph}.
Hence, the user needs foundation models and a model selection algorithm to handle such wide possibilities.
\begin{figure}[t!]
\centering
\begin{subfigure}[t]{0.5\textwidth}
    \centering
     \includegraphics[width=0.90\textwidth]{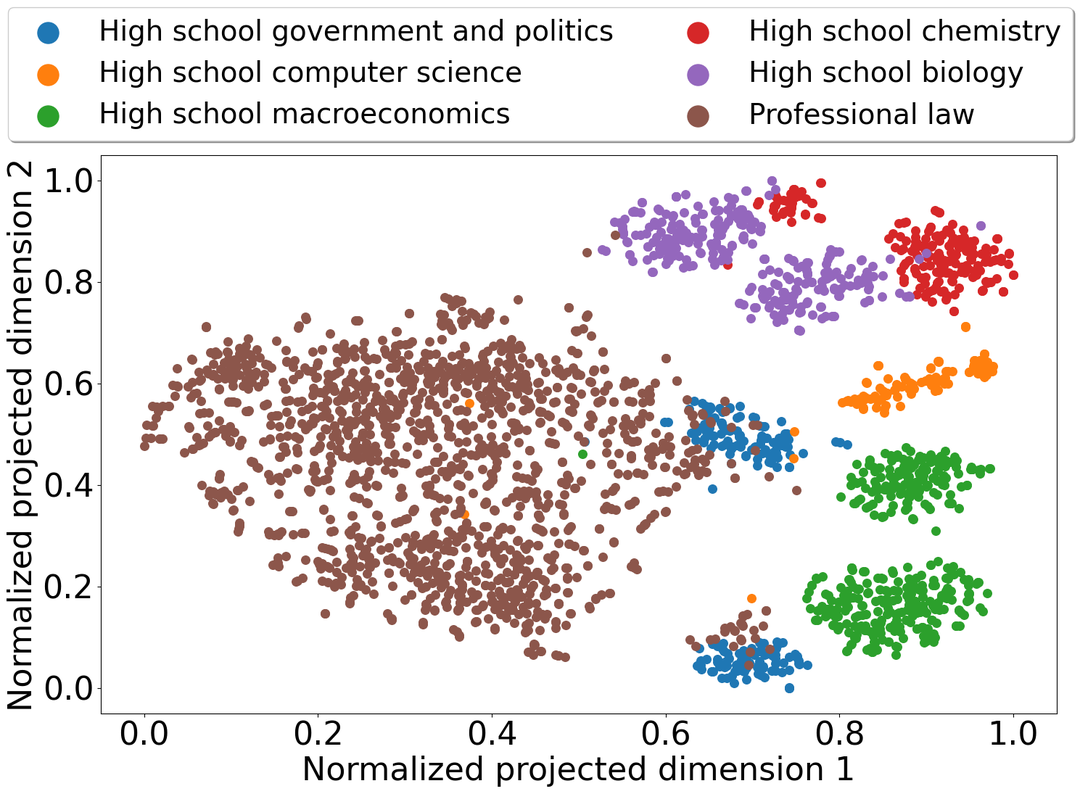} %
    \caption{\small{t-SNE visualization of MMLU data points.}}
    \label{fig:subtsne_mmlu}
\end{subfigure}
\hfill
\begin{subfigure}[t]{0.5\textwidth}
    \centering
    \includegraphics[width=0.93\textwidth]{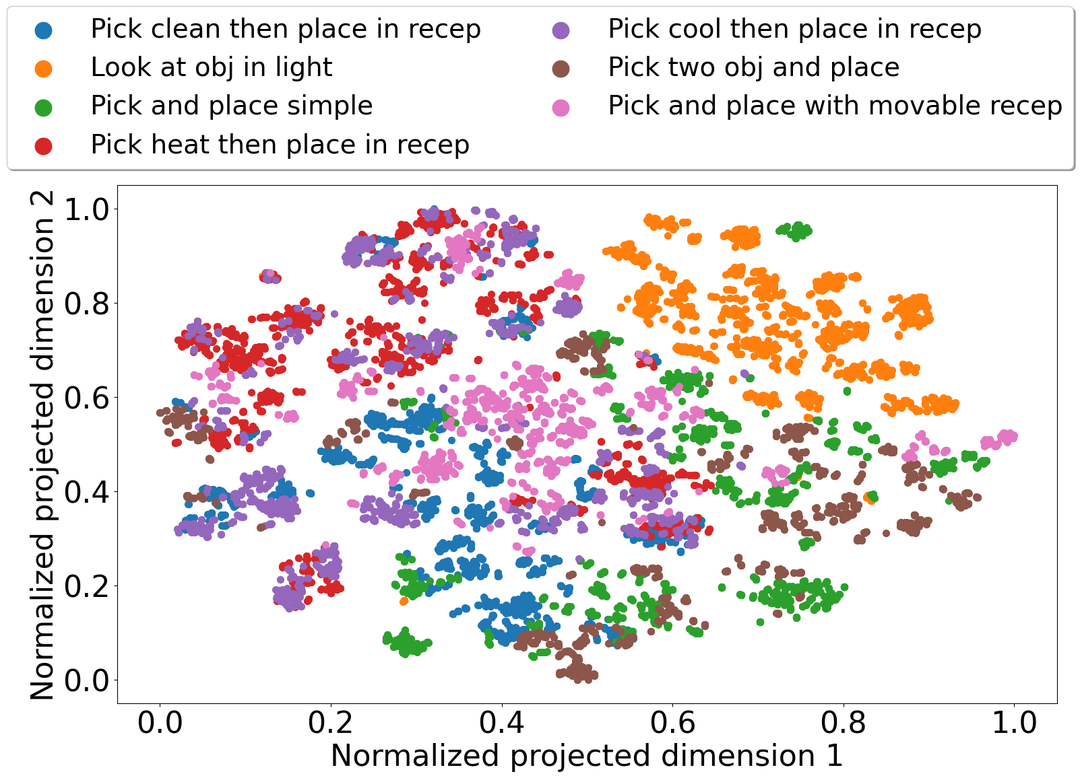}
    \caption{\small{t-SNE visualization of ALFRED instructions data points.}}
    \label{fig:subtsne_alfred}
\end{subfigure}
  \caption{\small{\textbf{Context naturally forms clusters:}
  The t-SNE visualization projects high-dimensional CLIP extracted features, \textit{i.e.}, the context, into $2$ dimensions, revealing feature clusters that correspond to categories of tasks, even without the use of such labels.
    }}
  \label{fig:tsne}
\vspace{-2em}
\end{figure}

\textbf{Feature extractor:} 
Given the wide range of data distributions, it becomes essential to employ a pre-trained encoder for feature extraction. 
It should be an open-source pre-trained foundation model \cite{clip, ViT, devlin2019bert, siglip} to ensure free user access.
The advantages of a pre-trained, open-source encoder are threefold: First, it can handle various data distributions, efficiently extracting features from user-provided data.
Second, the pre-trained model does not need additional training, enhancing data efficiency for our online model selection pipeline. 
One only needs data to train the model selection algorithm.
Third, the fixed-dimensional outputs of the encoder ensure smooth integration with all contextual online learning algorithms, which take fixed-length vectors as input. 
This distinct characteristic differs from transformer-like models, often used for natural language processing, which lack fixed-length output. Hence, the encoder is a better choice for our setting.
The online learning algorithm $\algo$ defined previously then takes the extracted feature $\cont_t$, \textit{i.e.}, the context, as input.

Ideally, the encoder clusters similar data, leading to close proximity between similar tasks.
This clustering enables contextual online learning algorithms to predict the expected performance and execution time of models for similar tasks, where the models demonstrate similar performance.
We use CLIP \cite{clip} encoder models to encode text and images into fixed-dimensional features.
We visualize the clustering of the features in MMLU and ALFRED in \reffig{fig:tsne}. 
MMLU, a question-answering dataset, and ALFRED, a language-based household robot dataset, are both detailed later.
We employ t-SNE \cite{t_sne} to reduce the CLIP features from $768$ to $2$ dimensions for better visual comprehension.
Different colors represent different tasks, demonstrating the encoder's ability to cluster data from tasks.

\textbf{Local open-source and remote closed-source models:} 
The user can select to execute various models. One of these models is open-source and local with free access. It is smaller in size and less powerful.
Remote closed-source models operate as paid services, incurring costs for users. These remote models are typically larger and offer higher performance, each with its unique attributes and capabilities. The local model offers users a cost-effective alternative, allowing them to execute it without paying. Ideally, users should use the local model for easier tasks and turn to the remote ones for difficult tasks that the local model cannot handle.

\begin{figure}[t!]
\centering
\begin{subfigure}[t]{0.5\textwidth}
    \centering
     \includegraphics[width=0.93\textwidth]{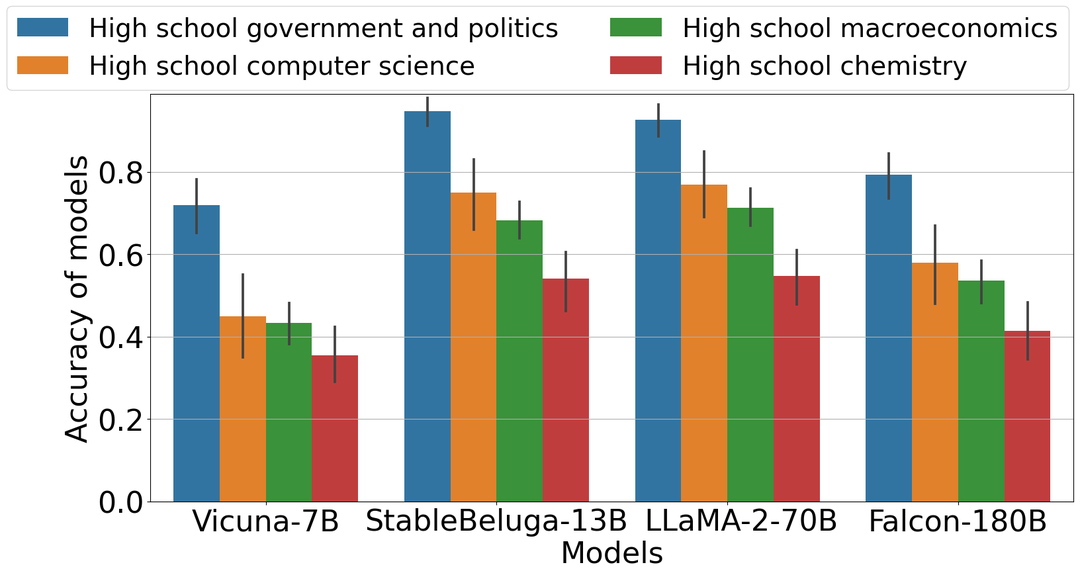} %
    \caption{\small{MMLU}}
    \label{fig:subtask_mmlu}
\end{subfigure}
\hfill
\begin{subfigure}[t]{0.5\textwidth}
    \centering
    \includegraphics[width=0.93\textwidth]{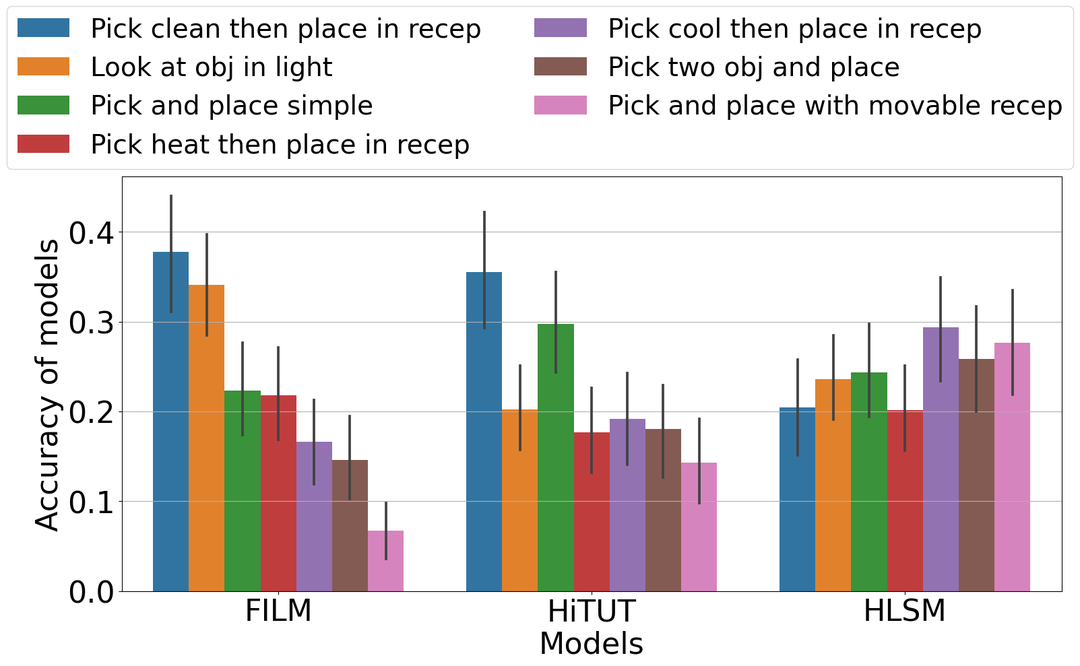}
    \caption{\small{ALFRED}}
    \label{fig:subtask_alfred}
\end{subfigure}
  \caption{\small{\textbf{Models perform differently on various tasks:} 
    Each model exhibits distinct performance on various tasks, colored differently. 
     The models on the left are the local models with the overall lowest performance.
     This variation in performance highlights the need to select the most effective models for each task or even data point. The error bars in the plots represent the $95\%$ confidence intervals.
    }}
    \label{fig:subtask}
\vspace{-2em}
\end{figure}

\begin{figure}[t!]
\centering
\begin{subfigure}[t]{0.5\textwidth}
    \centering
     \includegraphics[width=0.93\textwidth]{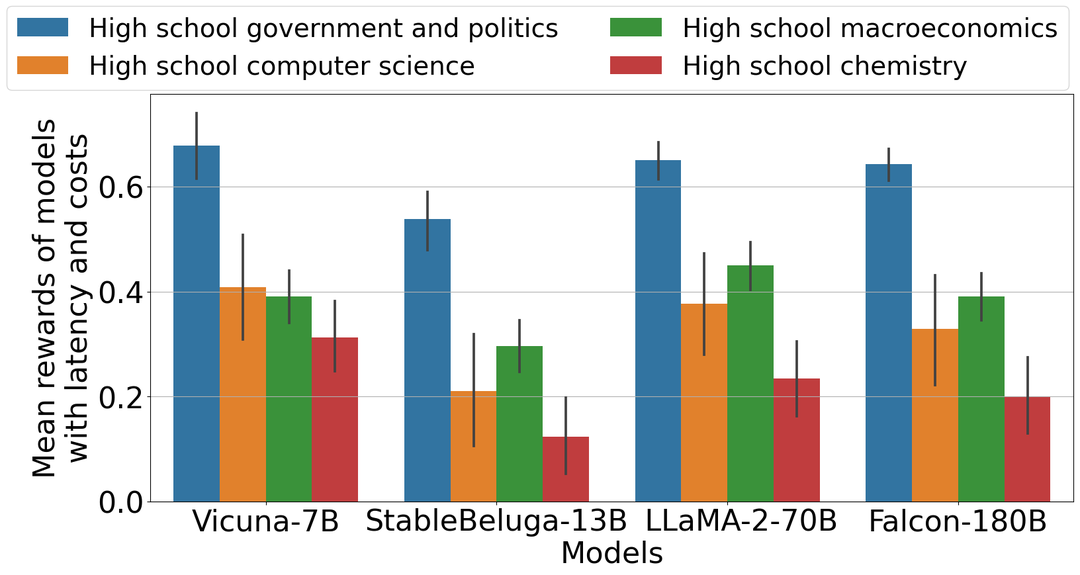} %
    \caption{\small{MMLU}}
\end{subfigure}
\hfill
\begin{subfigure}[t]{0.5\textwidth}
    \centering
    \includegraphics[width=0.96\textwidth]{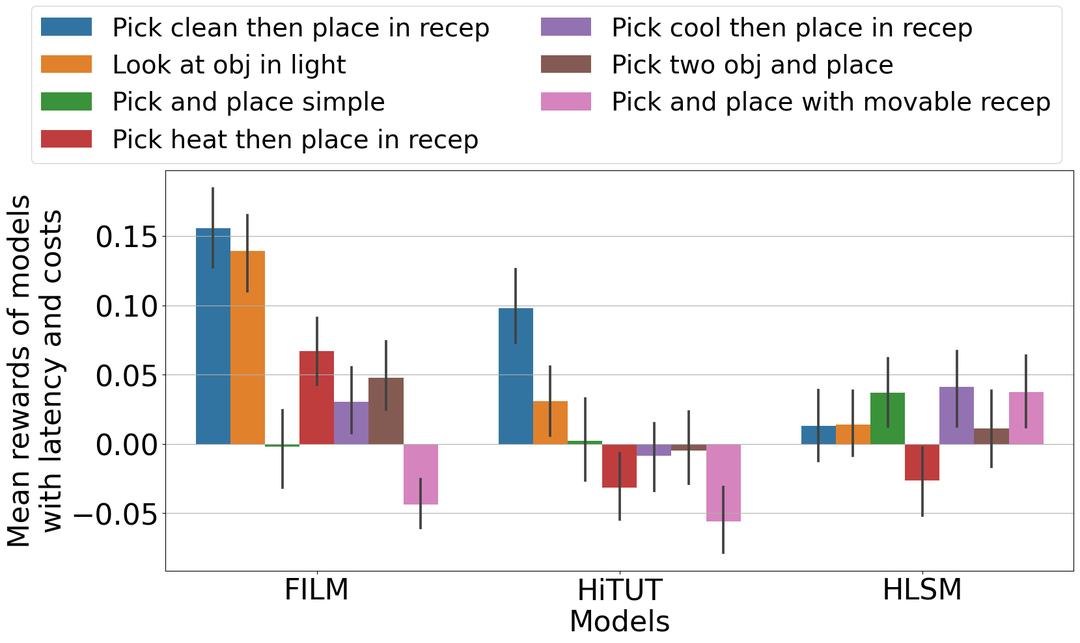}
    \caption{\small{ALFRED}}
\end{subfigure}
  \caption{\small{\textbf{Models perform differently on various tasks with latency and costs:} 
    Each model exhibits distinct performance on various tasks, represented by different colors. The models on the left are the local models. With latency and costs, the local model has no worse performance, as it is faster and free to execute.
    }}
  \label{fig:subtask_latency}
\vspace{-2em}
\end{figure}

We show the performance of the models across tasks in \reffig{fig:subtask}. While larger models have more capabilities, there are instances where a smaller model might be more suitable. 
In particular, for macroeconomics and computer science, LLaMA-2, with $70$ billion parameters, is a better choice than Falcon, with $180$ billion parameters. 
A similar trend exists in \reffig{fig:subtask_alfred} though the comparison of model sizes is more complex. The models in \reffig{fig:subtask_alfred} are combinations of memory storage units and neural networks, highlighting the importance of evaluating models based on their final performance (rewards), rather than solely on the parameters of the neural networks.
This observation challenges the simple approach of always selecting the largest model regardless of costs.
While in \reffig{fig:subtask} the local models on the left exhibit the overall lowest performance, this trend does not extend to scenarios with latency and costs caused by model execution in \reffig{fig:subtask_latency}. The local models demonstrate advantages over the remote and closed-source counterparts due to their fast execution and free access. 
It indicates that factors beyond model performance, such as execution time and monetary costs, make the model selection problem more complicated than simply selecting powerful close-source models.
Note that the rewards may become negative as the execution time and monetary costs penalize the rewards. 
However, they have no impact on our formulation since \refprob{prob} does not specify the range of rewards.

\textbf{Online model selection algorithm:}
Since obtaining complete knowledge of the capabilities of the models is impossible, we rely on a learning algorithm to explore and exploit their capabilities.  
We employ a contextual online model selection algorithm to learn the most suitable model given features extracted from a data point in real time.
The algorithm deduces the relationship between the features and rewards, representing the user's satisfaction with the model performance. Only the selected model reveals its reward, while the rewards of the other models remain concealed.
From \reffig{fig:tsne} and \reffig{fig:subtask}, it is evident that the features of the same task naturally form distinct clusters and that the models demonstrate proficiency in various tasks. Intuitively, a model selection algorithm first categorizes the cluster of features and then identifies the optimal model for each cluster by online exploration and exploitation.
When a new model becomes available, the algorithm can add it to the set $\mathcal{F}$ of models and adaptively continue learning. This adaptability is crucial \textit{given the constant emergence of new closed-source models from organizations around the world}. 

\textbf{Rewards:}
The reward is a scalar metric that measures the user's satisfaction with the outcome of the selected model. The objective of the model selection algorithm is to maximize this reward, thereby continuously enhancing the user's overall satisfaction.
There are $3$ terms in \refeq{eq:reward} that affect the user's satisfaction: model performance, execution time, and monetary costs. The definition of model performance relies on the specific task the model addresses, such as correctness in question answering or the success of tasks a household robot.
The execution time contains various components.
when opting for a closed-source model running on a remote private server, it contains the data transmission time from the local device to the server, the model execution time at the server, and the transmission time of the model's outputs back to the local device.
In contrast, selecting a local model involves only its execution time. Also, for certain tasks involving long-horizon planning, the execution time involves the steps taken by robots.
The monetary costs are the expenses incurred for closed-source models, which differ from service providers. These costs are in various units, such as per input and output language token or per action performed by a robot.
The $\alpha$ coefficients in \refeq{eq:reward} balance the $3$ factors, representing the user's preference among these terms.

\textbf{Online learning process:}
Since online learning falls into the broader category of reinforcement learning (RL), we can view \refprob{prob} as an RL problem. The action space of the RL problem corresponds to the set $\mathcal{F}$ of models, the context $\boldsymbol{g}(\x_t)$ serves as the observed state, and the reward remains the reward. 
The discount factor $\gamma$ of the problem is $0$ as our online model selection problem is not a sequential decision-making process. The previous states and the future ones are not causal. That is, the action taken at each time step does not affect future states, since the input data stream $\x_{1:T}$ is independent of the selected models $f^{a_t}$. 
Thus, we can use \textit{any on-policy RL algorithm} to solve \refprob{prob}, as they learn from the actions and rewards generated by the current policy. 
For the following, we employ a widely used on-policy algorithm--proximal policy optimization (PPO) \cite{schulman2017ppo}. Note that any on-policy algorithm would work, just as any open-source encoder would work in our pipeline.

\section{Analysis of Online Model Selection}
We now show that contextual online learning algorithms outperform non-contextual ones and analyze the impact of the non-negative weights $\alpha_{\tau}, \alpha_{\price}$ of latency and cost penalties on the online model selection problem.
Intuitively, the data points naturally form clusters within the encoder's feature space, as shown in \reffig{fig:tsne}. Contextual algorithms select models based on the cluster in which a data point is located. In contrast, non-contextual algorithms learn to select the overall best model, regardless the context or data clusters.

In the following analysis, we consider the case where $T \rightarrow \infty$, \textit{i.e.}, the online learning algorithm optimizes for an infinite horizon. We thus focus on the case where the online learning algorithm converges, thus acting optimally based on the available contextual information. The rewards from the early exploration phase are negligible due to the infinite horizon.
We simplify our setting to the case where there is only $1$ local model and $1$ remote model and assume that the encoder clusters the data points into $\Ncluster$ clusters, ${\cluster^1, \cluster^2, ..., \cluster^\Ncluster}$. 
We want to know the proportion of data points in that cluster that either fail or succeed in the two models, which we call the \textit{outcome rate}.
We denote this rate as $\suc_{S,F}$, where the first subscript indicates failure (F) or success (S) in the local model, and the second corresponds to the remote model.
We show the outcome rates for the overall datasets $\suc^{\mathrm{all}}$ in Table \ref{table:accuracy}. 
We then define the outcome rates and the proportion of data points for each cluster $i$ as $\suc^{i}$ and $\propor^{i}$ for all $i=[1, \Ncluster]$. 
We know from the definition of probability that:
$$\suc^{i}_{S,S} + \suc^{i}_{S,F} + \suc^{i}_{F,S} + \suc^{i}_{F,F} = 1 ~~ (\text{outcome rates sum to }1),$$ 
$$\sum_{i=1}^{\Ncluster} \propor^{i} = 1 ~~ (\text{proportions of data sum to }1), \quad \forall i \in [1, \Ncluster].$$ 
We further define $a=0$ as the action of selecting the local model and $a=1$ as the remote model.

We start with the analysis of a non-contextual online learning algorithm, which has no encoder.
Its converged action is simple, which is
\begin{equation}
a^{\mathrm{all}} = \boldsymbol{\mathbbm{1}}( 1-\suc^{\mathrm{all}}_{F,S} < \suc^{\mathrm{all}}_{F,S} - \alpha_{\tau} \tau^{\mathrm{all}} - \alpha_{\price} \price^{\mathrm{all}} ),
\label{eq:action_all}
\end{equation}
where $\boldsymbol{\mathbbm{1}}$ is the indicator function, and $\tau^{\mathrm{all}}, \price^{\mathrm{all}}$ are the mean execution time and costs over the data distribution. 
We rewrite $\suc^{\mathrm{all}}_{S,F}+\suc^{\mathrm{all}}_{S,S}+\suc^{\mathrm{all}}_{F,F}$ as $1-\suc^{\mathrm{all}}_{F,S}$ to simplify our notation.
Intuitively, it selects the remote model if the net gain from selecting it exceeds the benefits of selecting the local model. 
The benefit is the proportion of data that only the remote model can correctly process minus any penalties for selecting.
Such an algorithm naïvely selects the same model all the time, resulting in the mean cumulative reward:
\begin{equation}
    \frac{1}{T} \mathbb{E}^{\mathrm{all}}_{f^{a_t}, \mathcal{X}} \left[ \sum_{t=1}^T \reward_t \right] = \max \biggl( 1-\suc^{\mathrm{all}}_{F,S}, ~ \suc^{\mathrm{all}}_{F,S} - \alpha_{\tau} \tau^{\mathrm{all}} - \alpha_{\price} \price^{\mathrm{all}} \biggr).
    \label{eq:reward_all}
\end{equation}
\citet{ghosh2022dynamic} conducted similar analyses as well.

\begin{table}[t!]
\centering
\begin{tabular}{|| c || c c ||} 
 \hline
 \textbf{MMLU} & \multicolumn{2}{c||}{Remote models} \\ [0.5ex] 
  \hline 
 Local model & Success & Failure \\ 
 \hline 
 Success & $\suc^{\mathrm{all}}_{S,S}$ = 59\% & $\suc^{\mathrm{all}}_{S,F}$ = 1\% \\ 
 Failure & $\suc^{\mathrm{all}}_{F,S}$ = 27\% & $\suc^{\mathrm{all}}_{F,F}$ = 13\% \\ [1ex] 
 \hline
\end{tabular}

\vspace{1em}
\hspace{\fill}

\begin{tabular}{|| c || c c ||} 
 \hline
 \textbf{ALFRED} & \multicolumn{2}{c||}{Remote models} \\ [0.5ex] \hline 
 Local model & Success & Failure \\
 \hline 
 Success & $\suc^{\mathrm{all}}_{S,S}$ = 12\% & $\suc^{\mathrm{all}}_{S,F}$ = 11\% \\ 
 Failure & $\suc^{\mathrm{all}}_{F,S}$ = 29\% & $\suc^{\mathrm{all}}_{F,F}$ = 48\% \\ [1ex] 
 \hline
\end{tabular}
\caption{\small \textbf{Outcome rates of models:}
    The tables compare the performance of local and remote models for MMLU and ALFRED datasets. Each cell represents the proportion of data in different combinations of successes and failures between the local model and remote models. 
    If any one of the remote models successfully executes the task, it counts as a success here. Hence, the remote models are more powerful and have higher proportions of success.
}
\vspace{-2.5em}
\label{table:accuracy}
\end{table}

We now extend our analysis to clusters of data, thus being contextual. A contextual online learning algorithm outputs specific actions for each cluster. Our analysis incorporates the proportions of data within clusters $\propor^i$. Similar to \refeq{eq:action_all}, the optimal action $a^{i}$ of cluster $i$ is to compare the performance of local and remote models in cluster $i$: 
\begin{equation}
    a^{i} = \boldsymbol{\mathbbm{1}} \biggl(1-\suc^{i}_{F,S} < \suc^{i}_{F,S} - \alpha_{\tau} \tau^{i} - \alpha_{\price} \price^{i} \biggr).
    \label{eq:action_cluster}
\end{equation}
Then, we consider the expected reward for each cluster $i$:
\begin{equation}
   \mathbb{E}_{f^{a_t}, \mathcal{X}}^i \left[ \reward \right] = \max \biggl( 1-\suc^{i}_{F,S}, \suc^{i}_{F,S} - \alpha_{\tau} \tau^{i} - \alpha_{\price} \price^{i} \biggr).
\label{eq:reward_cluster}
\end{equation} 
This reward results from the previous actions in \refeq{eq:action_cluster} just like \refeq{eq:reward_all}.
Finally, we obtain the overall cumulative mean reward, which is a weighted sum of the cumulative mean reward of each cluster:
\begin{equation}
    \frac{1}{T} \mathbb{E}^{\mathrm{context}}_{f^{a_t}, \mathcal{X}} \left[ \sum_{t=1}^T \reward_t \right] = 
     \sum_{i=1}^{\Ncluster} \propor^{i} \mathbb{E}_{f^{a_t}, \mathcal{X}}^i \left[ \reward \right].
    \label{eq:reward_all_cluster}
\end{equation}

The non-contextual algorithm selects the remote model when $\suc^{\mathrm{all}}_{F,S}>0.5$, as indicated by \refeq{eq:action_all} equals $1$. This naïve outcome reveals the flaws of the non-contextual algorithm. It only selects the overall best model, disregarding the contextual information in the data. 
In contrast, the contextual algorithm selects the best model based on the outcome rate of each cluster of data, making it better. Comparing \refeq{eq:reward_all} and \refeq{eq:reward_all_cluster}, we can see this superiority: 
\begin{equation}
\begin{aligned}
    \frac{1}{T} \mathbb{E}^{\mathrm{context}}_{f^{a_{t}}, \mathcal{X}} & 
    \left[ \sum_{t=1}^{T} \reward_{t} \right] 
    \geq 
    \frac{1}{T} \mathbb{E}^{\mathrm{all}}_{f^{a_t}, \mathcal{X}} \left[ \sum_{t=1}^T \reward_{t} \right] \\
    & (\text{superiority of contextual algorithms})
\end{aligned}
\end{equation}
due to the convexity of the $\max$ function. 
From \refeq{eq:reward_cluster}, we can also see the effect of the encoder on the mean cumulative reward.
The optimal encoder should cluster data points that exhibit similar performance in one model but vary significantly in both models. 
For each cluster $i$ clustered by the encoder, the absolute difference
$|(1-\suc^{i}_{F,S}) - ( \suc^{i}_{F,S} - \alpha_{\tau} \tau^{i} - \alpha_{\price} \price^{i})|$ should be significant. 
It suggests that the clustered data points are exclusively suitable for the remote or local model. 

We now study the effects of the non-negative weights $\alpha_{\tau}, \alpha_{\price}$ of time and cost penalties. 
Intuitively, if the weights are large, the reward of selecting the remote model is low, and all algorithms learn to select the local model.
Precisely, a contextual algorithm always selects the local model if $\alpha_{\tau}, \alpha_{\price}$ satisfy one of the following conditions: 
\begin{equation}
\begin{aligned}
    a^i = 0, & ~~ \forall i \in [1, \Ncluster] \\
    \iff 1-\suc^{i}_{F,S} \geq \suc^{i}_{F,S} - \alpha_{\tau} \tau^{i} - \alpha_{\price} \price^{i}, & ~~ \forall i \in [1, \Ncluster].
\end{aligned}
\end{equation}
In contrast, it always selects the remote model when $\alpha_{\tau}, \alpha_{\price}$ satisfy one of the following conditions: 
\begin{equation}
\begin{aligned}
    a^i = 1, & ~~ \forall i \in [1, \Ncluster] \\
    \iff 1-\suc^{i}_{F,S} \leq \suc^{i}_{F,S} - \alpha_{\tau} \tau^{i} - \alpha_{\price} \price^{i}, & ~~ \forall i \in [1, \Ncluster].
\end{aligned}
\end{equation}
Therefore, our interest lies in cases other than the previous trivial ones, where the algorithm selects the remote or local model across clusters. Namely, when $\alpha_{\tau}, \alpha_{\price}$ satisfy
\begin{equation}
\begin{aligned}
    \exists i, ~ \text{s.t. } & 1-\suc^{i}_{F,S} > \suc^{i}_{F,S} - \alpha_{\tau} \tau^{i} - \alpha_{\price} \price^{i} \\
    \wedge ~ \exists j, ~ \text{s.t. } & 1-\suc^{j}_{F,S} < \suc^{j}_{F,S} - \alpha_{\tau} \tau^{j} - \alpha_{\price} \price^{j}.
\end{aligned}
\end{equation}
Only in these non-trivial cases, the contextual learning algorithm balances the trade-off between the local and remote models, considering the specific costs and benefits associated with each cluster.
One can extend the previous analysis to settings of multiple remote models by conceptually merging all remote models into one. This conceptual model internally selects the most suitable remote model for each data point. Then, the subsequent steps adhere to our previous analysis.

\section{Experiments}
\subsection{Settings:}
We tested our pipeline on testing sets from $4$ different tasks:
(\textit{a}) Massive Multitask Language Understanding (MMLU) question answering \cite{hendryckstest2021mmlu}, (\textit{b}) Waymo Open Dataset \cite{Sun_2020_Waymo_dataset} multi-modal object querying, (\textit{c}) ALFRED \cite{ALFRED20}, a simulated environment for robots to interpret and perform everyday tasks, and (\textit{d}) Open X-Embodiment \cite{RTX}, the largest open-source real robot dataset. It contains more than $1$ million real robot trajectories that span $22$ robot embodiments.
Due to constraints in resources and accessibility, we were unable to test our pipeline massively on real closed-source models. However, we conducted tests on models of varying sizes, simulating situations with one small local model and multiple large remote closed-source models.
Although MMLU is a dataset of language-based multi-disciplinary problems and is not related to robotics, we used it to illustrate our pipeline's capacity to orchestrate real-world large language foundation models. These models come from various organizations and address multi-disciplinary questions.
The Waymo Open Dataset showcases scenarios where users use language to ask an autonomous vehicle about the presence of specific objects in its visual observations.
In ALFRED and Open X-Embodiment, we demonstrated the application of language-based robotics in everyday tasks. They enable users to articulate their intentions using unstructured natural sentences.

\begin{figure*}[ht!]
  \centering
  \includegraphics[width=0.98\textwidth]{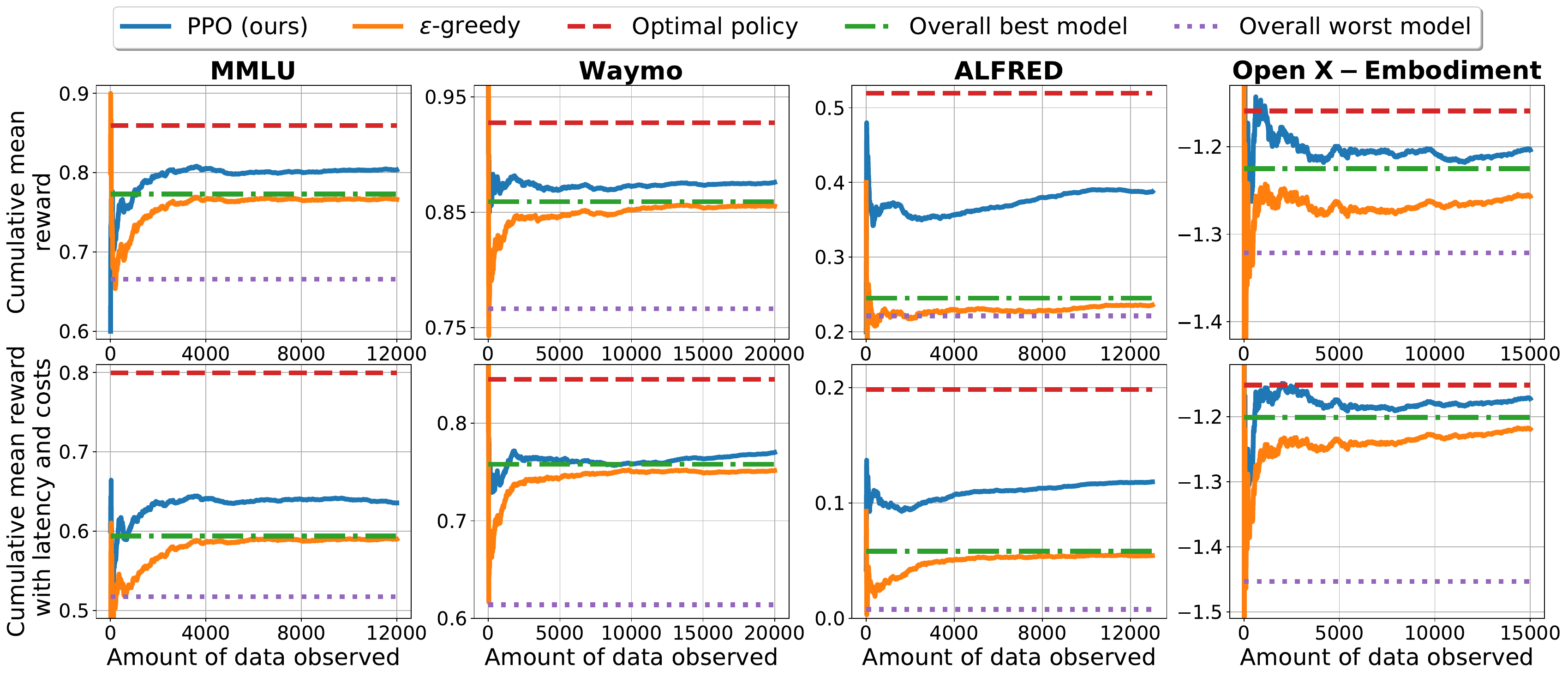} %
  \caption{\small{\textbf{Contextual PPO algorithm outperforms baselines in the online model selection problem:}
    Our contextual PPO algorithm outperforms non-contextual algorithms, represented as orange, green, and purple lines. Our agent achieves higher cumulative mean rewards with and without latency penalty. This underscores the importance of incorporating context into our model selection pipeline. We also provide the Oracle optimal policy as the ultimate upper bound for all benchmarks.
    }}
\vspace{-2em}
  \label{fig:bandit_results}
\end{figure*}

We chose the model with the lowest performance as the local model.
The model execution time is the sum of data transfer time and model execution time on GPU, measured in logarithmic scale ($\log_{10}$ seconds) for MMLU, Waymo, and Open X-Embodiment datasets. The model execution time is the robot's execution steps for ALFRED.
We based the monetary costs on per-token, as per GPT-4. 
We used an NVIDIA Jestom Orin as the local device and an NVIDIA RTX $6000$ Ada Generation $48$ GB as the remote device for the Waymo dataset, and we only used NVIDIA RTX $6000$ Ada Generation $48$ GB for other tasks, as it is impossible to run the models on Jetson Orin.
In \refapp{sec:network_stat}, we presented statistics on network transmission and model execution time. We measured the data transmission time by connecting a Jetson Orin to servers in the backbone network via cable ethernet.
We tested all models on the available testing sets, ensuring that all testing data observed by each model are uniformly sampled without replacement.
See the example of prompts and images in \reffig{fig:datasets} and the details of the datasets in \refapp{sec:datasets}.

\textbf{MMLU:}
We use $4$ language models, Vicuna-$7$B \cite{vicuna}, StableBeluga-$13$B \cite{mukherjee2023orca, touvron2023llama2}, LLaMA-$2$-$70$B \cite{touvron2023llama2}, and Falcon-$180$B \cite{falcon}, to answer MMLU, a dataset covers multiple-choice questions from $57$ disciplinaries including computer science, law, etc. Each data point contains a question and $4$ choices, where only one is correct.
These models of various sizes, as indicated by the numbers of billions of parameters in the names, try to answer the prompted questions. The performance of the models is the correctness of the selected answers.
We intentionally choose models of various sizes to demonstrate the trade-off between model performance and execution time, as larger models are more accurate but slower to execute.

\textbf{Waymo Open Dataset:}
Waymo Open Dataset consists of hours of ego-centric self-driving car images. We fed these images along with prompts asking whether an object exists in the image to LLaVA \cite{liu2023improvedllava} models. 
These LLaVA models include LLaVA-$1.5$-$7$B, LLaVA-$1.5$-$13$B, LLaVA-$1.5$-$13$B-LoRA, which differ in training sets and sizes, from $7$ to $13$ billion parameters.
These models process images with corresponding language prompts and output text descriptions. We evaluated the performance of the models by their accuracy in identifying $10$ specific objects, such as trucks and cars, from the images.
The binary score reflecting the correctness of identifying these objects indicates the model performance.

\textbf{ALFRED:}
ALFRED is a simulated environment of a household robot that interacts with objects in a room. It consists of thousands of high-level human language instructions and the robot's actions to perform household tasks in rooms.
The high-level instructions provide an overall task description, as in \reffig{fig:datasets} and \reffig{fig:system_graph}. Each task consists of subgoals that form a sequence of goals to achieve the overall objective.
The dataset also includes floor plans for each room. We tested $3$ state-of-the-art models of ALFRED \cite{zhang-chai-2021-hitut, blukis2022hlsm, min2021film}, which process high-level instructions and the robot's ego-centric view to determine its actions. The selected model operates until the task succeeds, fails, or reaches $1,000$ steps. The model performance is a binary success indicator score. Note that the models exhibit low performance in this dataset, as long-horizon language-based planning still remains a challenge in robotics.

\textbf{Open X-Embodiment:}
Open X-Embodiment, unlike the simulated environment of ALFRED, is a supervised-learning dataset containing video demonstrations of robots with ground truth actions. 
Models trained on this dataset learn to infer actions from the observed images and the natural language intentions. We tested the dataset on two models--Octo-small and Octo-base \cite{octo_2023}. We used the negative mean squared error of inferred actions and ground truth actions as the model performance, as we do not have the original environments to evaluate the success rate of the task.

\textbf{Baselines:}
Throughout all experiments, we used PPO \cite{schulman2017ppo} as the contextual online learning algorithm.
We compared it with the best and worst single models and a non-contextual online learning algorithm, the $\epsilon$-Greedy algorithm \cite{BanditAlgorithms}.
We calculated the overall best and worst models based on their averaged performance across the testing sets. The $\epsilon$-Greedy algorithm has an increasing probability of selecting the empirically best model in the process of exploring alternative models.
Since it is non-contextual, the best strategy it can learn is to always select the top-performing model. Therefore, its performance has an upper bound set by the best single model, which is the overall best model in \reffig{fig:bandit_results}.
We also compared with the optimal policy, which reflects the Oracle case where one has complete knowledge of all models' execution results, enabling it to select the model maximizing the reward for every data point. 
The optimal policy does not always guarantee success, as there are instances in which all models fail. %

\subsection{Results:}
\textbf{Can we learn to select models on the fly?}
We showed the effectiveness of PPO in \reffig{fig:bandit_results}.
PPO obtained a higher cumulative reward than the best model, demonstrating the advantage of model selection. 
It implies the capability of selecting the most suitable model for each specific data point rather than defaulting to the overall best model. 
Since users are unable to determine the overall best or worst model without accessing the whole testing set, the PPO's ability to outperform the overall best model with limited data highlights the data-efficiency of our pipeline in model selection.
\refapp{sec:hyperpara} gives the hyperparameters of all algorithms.
We showed the online model selection problem with model performance rewards only ($\alpha_{\tau}=\alpha_{\price}=0$) in the top row of \reffig{fig:bandit_results}. The bottom row of \reffig{fig:bandit_results} addresses the case with execution time and monetary costs. 
We can see that in all experiments, PPO successfully learned to maximize rewards, outperforming the performance of the best single model. 

\textbf{How fast can we learn a contextual algorithm?}
At the beginning of the learning process, stochasticity heavily influenced the cumulative mean reward $R_{\mathrm{cum}}$.
For example, the algorithm might luckily select models that correctly perform the task multiple times consecutively early on, resulting in a reward exceeding the optimal policy.
Over time, the law of large numbers stabilized the rewards. 
Notably, PPO converged when observing thousands of data points, significantly fewer than what is usually required for supervised-learning datasets.
It emphasizes the benefits of integrating an open-source encoder into our framework to accelerate learning.
Otherwise, one needs extensive data to train an encoder first and then the PPO algorithm. 
Alternatively, bypassing the encoder requires training a PPO algorithm capable of processing both images and texts, which again demands significant data for training.
We also tried other contextual bandit algorithms adapted from non-contextual ones, as per \cite{cortes2018contextual_bandits}. It did not show a significant improvement in cumulative mean reward $R_{\mathrm{cum}}$ and requires more training time.
These algorithms assume binary rewards and do not align with our reward function with latency and costs, so we do not show their results here.

\begin{figure}[t!]
\centering
\includegraphics[width=0.47\textwidth]{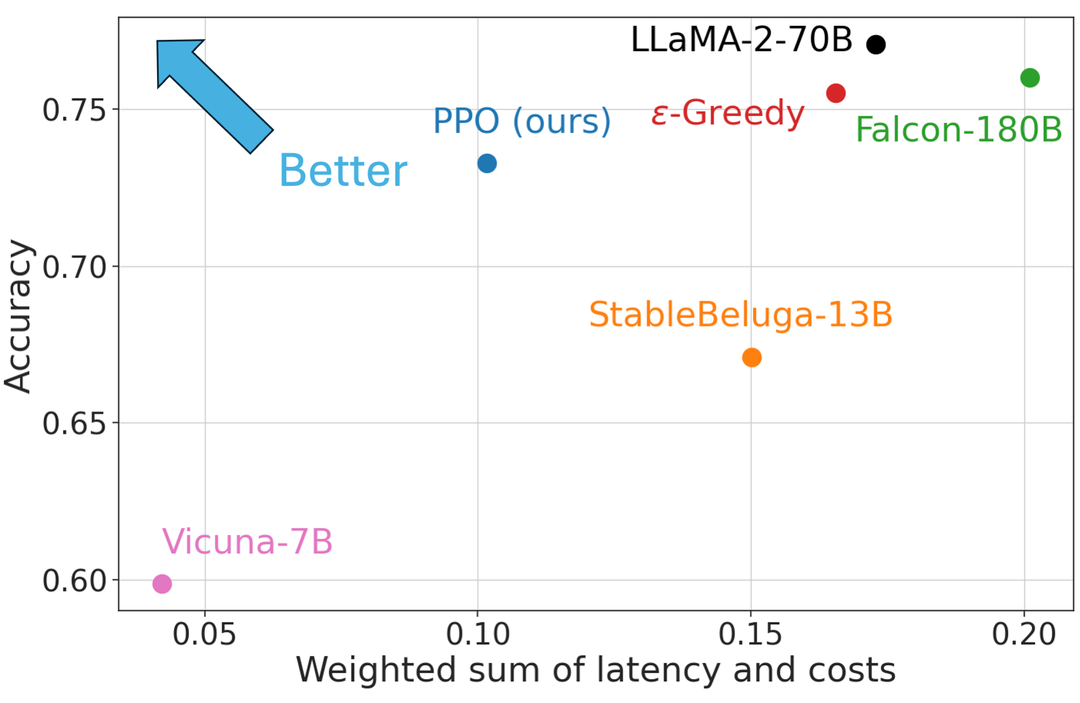}
    \caption{\small{\textbf{Model selection algorithm leads to a better trade-off between accuracy, latency, and costs:} 
    Our contextual PPO algorithm balances model accuracy with latency and costs, leading to a maximum mean reward compared to other benchmarks. We used the MMLU dataset as an example here.
    }}
\label{fig:acc_time}
\vspace{-2em}
\end{figure}

\textbf{Does PPO reach a better trade-off between accuracy, latency, and costs?}
\reffig{fig:acc_time} shows a comparison of the accuracy of various baseline methods and the weighted sum of latency and costs. The contextual PPO algorithm achieved higher accuracy while maintaining lower latency and costs, as indicated by its position towards the upper left. It outperformed other baseline methods, suggesting a more efficient trade-off between accuracy, latency, and costs. Also, the $\epsilon$-Greedy algorithm is close to the best model in \reffig{fig:acc_time} since it gradually learns to select only the best model, leading to similar performance.

\subsection{Limitations:}
Our pipeline is a heuristic to the online model selection problem, lacking any theoretical guarantees on the convergence rate or cumulative rewards compared to the optimal policy. 
Finding the best set of hyperparameters of the online learning algorithm requires data, which is not considered here. We only show the results of the agents with tuned hyperparameters.
Additionally, our pipeline cannot utilize any prior knowledge users may have about the models, which could otherwise speed up the learning process.
Lastly, if a model significantly outperforms and operates faster than all others across all data distributions, then the optimal policy would be trivial and straightforward--always selecting this ideal model. 
However, it is uncommon as powerful models tend to be larger and thus slower. This inherent trade-off between model performance and execution time sophisticates the model selection problem.

\section{Conclusion and Future Work}
We formulated the closed-source model selection problem as a contextual online learning problem and thus introduced a new pipeline to the problem.
It uses context derived from user data to speed up learning algorithms.
We examined an online learning algorithm, PPO, to effectively solve this problem. 
We also showed the superiority of the contextual algorithm through theoretical analysis.
Our methods have proven effective, achieving up to a $14\%$ improvement in various language-based robotic tasks compared to non-contextual methods.

Future research directions involve cases where contextual bandit algorithms can choose several bandits simultaneously, merging their outputs to produce a more comprehensive action. Another direction is where multiple users can exchange their model selection outcomes to facilitate faster collaborative learning. It is similar to federated learning, where collaborative information exchange improves the learning process \cite{datagame, FleetRoboticData, FederatedLearningRobotics, konečný2017federated}.
Furthermore, pre-processing of multi-model features between multiple encoders could lead to better performance or training efficiency of the online learning algorithm \cite{li2023taskaware, norelli2023asif}.
Lastly, establishing theoretical guarantees for non-compact contextual bandits remains an unresolved challenge in online learning.

\printbibliography

\clearpage
\appendix
\subsection{Execution Time of Model Selection}
\label{sec:network_stat}
\begin{figure}[h!]
\centering
\includegraphics[width=0.48\textwidth]{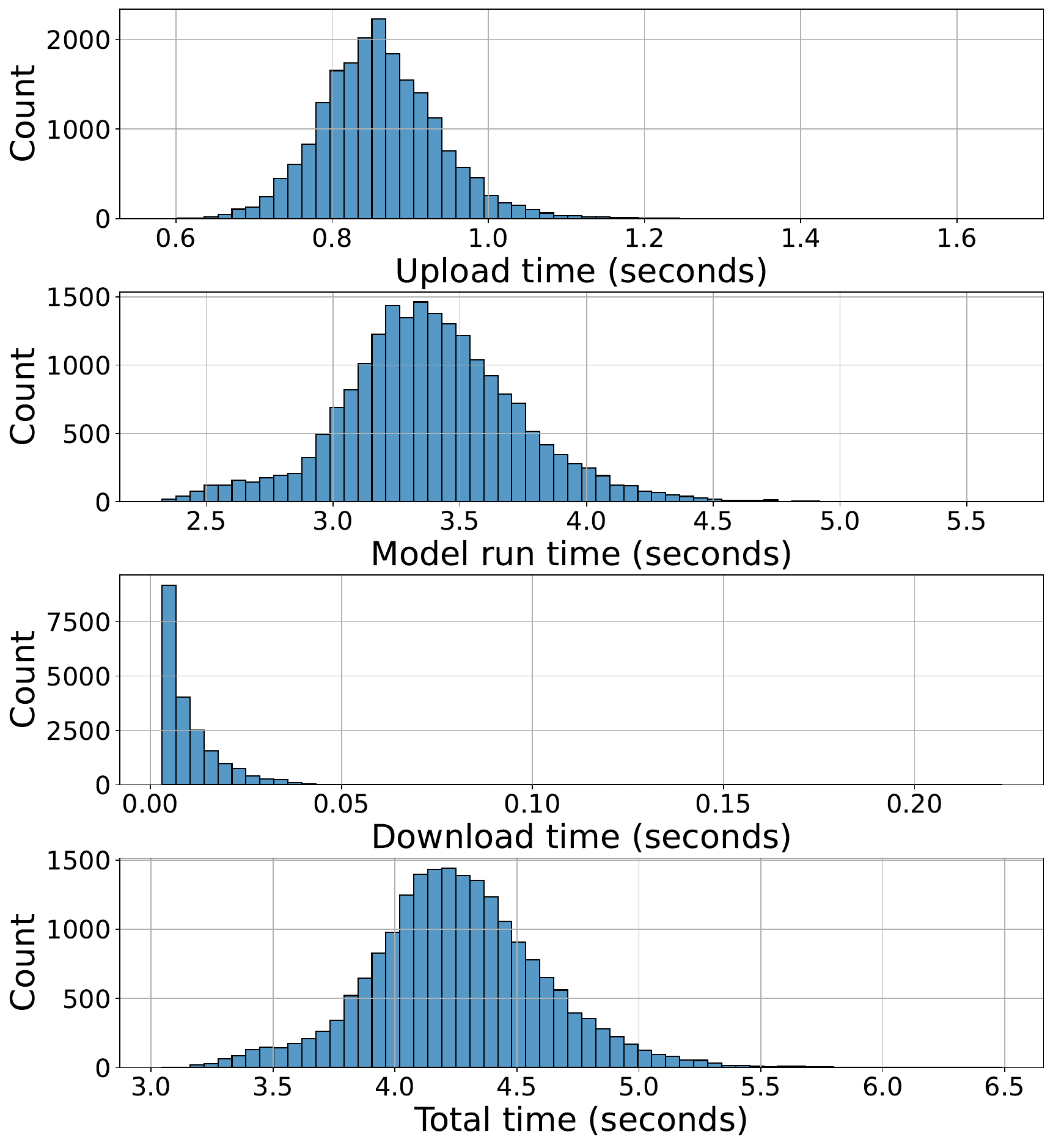}
\caption{\small{\textbf{Each phase of model execution time is not equally long.} 
    The plot shows the three phases of selecting remote models of the Waymo Open Dataset. The three phases are data uploading, model inference, and model output downloading. Each subplot presents the time distribution of a phase, with a total of $20,000$ data points collected.
}}
\label{fig:network_time}
\vspace{-1em}
\end{figure}

Selecting remote models involves three phases: (1) Uploading input data to remote servers where the models are located. (2) The inference time, \textit{i.e.}, the time it takes for the model to process the data. (3) Downloading the models' outputs back to the local device. We show the statistics of the three phases of the Waymo Open Dataset in \reffig{fig:network_time}.
We can see that the time spent uploading images and user querying texts accounts for about one-third of the model execution time.
The download time is considerably shorter in comparison to other phases because it involves transmitting back only the models' text responses, which are significantly smaller in size than uploaded images.

\subsection{Datasets}
\label{sec:datasets}

\textbf{Weights:}
We set the weights in \refeq{eq:reward} to $\alpha_{\tau}=0.03,\alpha_{\price}=0.0008$ for MMLU, $\alpha_{\tau}=0.2,\alpha_{\price}=0.001$ for the Waymo Open Dataset, $\alpha_{\tau}=0.05, \alpha_{\price}=0.005$ for ALFRED, and  $\alpha_{\tau}=0.2, \alpha_{\price}=0.01$ for the Open-X Embodiment Dataset.

\textbf{ALFRED:}
We use the seen and unseen validation scenarios of ALFRED datasets to test our pipeline. 
However, the validation set of the ALFRED dataset is not abundant, which merely has $1,600$ languages-instructed tasks. To augment the validation set, we use LLaMA-$2$ to paraphrase the human high-level instructions. Hence, we obtained roughly $13,0000$ languages-instructed data in total. 
For the experiments with latency and costs, we use the percentage of subgoals completed minus the execution time and financial penalty as the reward function.

\textbf{Waymo Open Dataset:}
We input the following prompt into the LLaVA models: \textit{"Answer whether or not there is a \{obj\} in the picture and where is it."}, where \textit{\{obj\}} is one of the following: [car, truck, bicycle, motorcycle, pedestrian, pole, sign, traffic light, building, vegetation].

\textbf{Open-X Embodiment:}
We randomly select the $15,000$ tasks from the bridge, kuka, and fractal sub-datasets to test our pipeline. We fed the last $2$ image observations of the task to the Octo models. 
For the contextual online learning algorithm, we fed the CLIP embeddings of the $2$ images and the embeddings of the language instructions into the model.

\subsection{Hyperparameters}
\label{sec:hyperpara}
For all experiments, we use the PPO implementation of Stable-Baselines3 \cite{stable-baselines3}, configuring the fully connected layers to the following dimensions: $[768, 512, 128, 64, 16]$. 
The hyperparameters of PPO include a learning rate of $0.0001$, a target KL divergence of $0.003$, and an entropy coefficient of $0.01$. Training occurs every $5$ steps, with other hyperparameters set to their default values. 

The exploration probability of the $\epsilon$ -Greedy algorithm at time $t$ ($t$-th step of action) is $$\epsilon = \min\{1, |\mathcal{F}| / ((t + 1) \times 0.01)\}.$$
We modified the exploration probability as per \cite{BanditAlgorithms}, where we chose the optimality gap of the arms as $0.01$.

\end{document}